\pdfoutput=1

\documentclass[11pt]{article}

\usepackage{emnlp2021}

\usepackage{times}
\usepackage{latexsym}

\usepackage[T1]{fontenc}

\usepackage[utf8]{inputenc}

\usepackage{microtype}

\usepackage{amsmath}
\usepackage{amssymb}
\usepackage[inline]{enumitem}
\usepackage[nameinlink,capitalise]{cleveref}
\usepackage{caption}
\usepackage{subcaption}
\usepackage{graphicx}
\usepackage{booktabs}
\usepackage{xcolor}

\crefformat{section}{\S #2#1#3}
\Crefformat{section}{\S #2#1#3}

\usepackage{hyperref}

\title{Continuous Entailment Patterns for Lexical Inference in Context}

\author{Martin Schmitt \and Hinrich Schütze \\
	Center for Information and Language Processing (CIS)\\
	LMU Munich, Germany\\
	\texttt{martin@cis.lmu.de}\\}

\newcommand{\conan}{\textsc{conan}}
\newcommand{\prem}{\mathbf{p}}
\newcommand{\hypo}{\mathbf{h}}
\newcommand{\lm}{\mathcal{M}}
\newcommand{\contokens}{C}
\newcommand{\auc}{\textsc{auc}}

\newcommand{\abs}[1]{\left|#1\right|}
\newcommand{\eset}[1]{\left\{\, #1 \,\right\}}
\newcommand{\floor}[1]{\left\lfloor #1 \right\rfloor}
\newcommand{\cprob}[2]{P\left( #1 \mid #2 \right)}

\newcommand{\para}[1]{\noindent\textbf{#1.}}

\begin{document}
\maketitle
\begin{abstract}
	Combining a pretrained language model (PLM) with textual patterns
	has been shown to help in both zero- and few-shot settings.
	For zero-shot performance, it makes sense to design patterns
	that closely resemble the text seen during self-supervised pretraining
	because the model has never seen anything else.
	Supervised training allows for more flexibility.
	If we allow for tokens outside the PLM's vocabulary,
	patterns can be adapted more flexibly to a PLM's idiosyncrasies.
	Contrasting patterns
	where a ``token'' can be any continuous vector vs.\ those
	where a discrete choice between vocabulary elements has to be made,
	we call our method \emph{CONtinuous pAtterNs} (\conan{}).
	We evaluate \conan{} on two established benchmarks
	for lexical inference in context (LIiC) a.k.a.\ predicate entailment,
	a challenging natural language understanding task with relatively small training sets.
	In a direct comparison with discrete patterns,
	\conan{} consistently leads to improved performance,
	setting a new state of the art.
	Our experiments give valuable insights into the kind of pattern that enhances a PLM's performance on LIiC
	and raise important questions regarding our understanding of PLMs using text patterns.\footnote{Our code is publicly available: \url{https://github.com/mnschmit/conan}}
\end{abstract}


\section{Introduction}

Lexical inference in context (LIiC) -- also called predicate entailment --
is a variant of natural language inference (NLI) or recognizing textual entailment \citep{dagan13}
with focus on the lexical semantics of verbs and verbal expressions \citep{levy-dagan-2016-annotating,schmitt-schutze-2019-sherliic}.
Its goal is to detect entailment between two very similar sentences,
i.e., sentences that share subject and object and only differ in the predicate,
e.g., \textsc{person}$(A)$ \emph{runs} \textsc{org}$(B)$ $\rightarrow$ \textsc{person}$(A)$ \emph{leads} \textsc{org}$(B)$.
%
%
%
NLI models that were not specifically trained with lexical knowledge have been reported to struggle with this task \citep{glockner-etal-2018-breaking,schmitt-schutze-2019-sherliic},
making LIiC an important evaluation criterion for general language understanding.
Other use cases for this kind of lexical entailment knowledge include question answering \citep{schoenmackers-etal-2010-learning,mckenna21}, event coreference \citep{shwartz-etal-2017-acquiring,meged-etal-2020-paraphrasing}, and link prediction in knowledge graphs \citep{hosseini-etal-2019-duality}.

Although LIiC is an inherently directional task,
symmetric cosine similarity in a vector space, such as word2vec \citep{mikolov13},
has long 
been the state of the art for this task.
Only recently transfer learning with pretrained Transformer \citep{vaswani17} language models \citep{devlin-etal-2019-bert},
has led to large improvements for LIiC.
\citet{schmitt-schutze-2021-language} combine natural language (NL) patterns with a pretrained language model (PLM)
and not only set a new state of the art but also beat baselines without access to such patterns.

Empirical findings suggest that a good pattern can be worth 100s of labeled training instances \citep{scao-rush-2021-many},
making pattern approaches interesting for low-resource tasks such as LIiC.
But beyond the intuition that patterns serve as some sort of task instruction \citep{schick-schutze-2021-exploiting},
little is known about the reasons for their success. 
Recent findings that
\begin{enumerate*}[label=(\roman*)]
	\item PLMs can fail to follow even simple instructions \citep{efrat20}, that
	\item PLMs can behave drastically different with paraphrases of the same pattern \citep{elazar21},
	and that
	\item performance increases if we train a second
	model to rewrite an input pattern
with the goal of making it more comprehensible for
a target PLM \citep{haviv-etal-2021-bertese},
\end{enumerate*}
strongly suggest that patterns do not make sense to PLMs in the same way as they do to humans.


Our work sheds light on the interaction of patterns and PLMs
and proposes a new method of improving pattern-based models fully automatically.
On two popular LIiC benchmarks,
our model 
\begin{enumerate*}[label=(\roman{*})]
	\item
establishes a new state of the art without the need for handcrafting patterns or automatically identifying them in a corpus and
	\item does so more efficiently thanks to shorter patterns.
\end{enumerate*}
Our best model only uses 2 tokens per pattern.



\section{The \conan{} Model}
\label{sec:conan}

\para{Continuous patterns}
LIiC
is a binary classification task,
i.e., given a premise $\prem = p_1p_2\dots p_{\abs{\prem}}$ and a hypothesis $\hypo = h_1h_2\dots h_{\abs{\hypo}}$ a model has to decide whether $\prem$ entails $\hypo$ ($y=1$) or not ($y=0$).
A template-based approach to this task surrounds $\prem$ and $\hypo$ with tokens $t_1t_2\dots t_{m}$ to bias the classifier for entailment detection, e.g., ``$\prem$, \emph{which means that} $\hypo$''.
While in most approaches
that leverage a
PLM $\lm$
these tokens come from the PLM's vocabulary, i.e.,
$\forall i.\,t_i\in\Sigma_{\lm}$,
we propose a model based on CONtinuous pAtterNs (\conan{}),
i.e., surround the embedding representation of $\prem$ and $\hypo$ with continuous vectors that may be close to but do not have to match the embedding of any vocabulary entry.

For this, we first extend the PLM's vocabulary by a finite set $\contokens = \eset{c_1, c_2, \dots, c_{\abs{C}}}$ of fresh tokens,
i.e., $\Sigma = \Sigma_{\lm} \cup \contokens$ with $\contokens \cap \Sigma_{\lm} = \emptyset$.
Then, we distinguish two methods of surrounding $\prem$ and $\hypo$ with these special tokens:
$\alpha$ sets them both $\alpha$round and between $\prem$ and $\hypo$ (\cref{eq:alpha}) while $\beta$ only sets them $\beta$etween the two input sentences (\cref{eq:beta}).
\begin{align}
	\label{eq:alpha}\alpha_k(\prem, \hypo) &= c_1\dots c_{a} \prem c_{a+1}\dots c_{a+b} \hypo c_{a+b+1}\dots c_k \nonumber\\
	\text{with } a = &\floor{k/3}, \hspace{.1em} b = \floor{k/3} + (k \bmod 3) \\[.75em]
	\label{eq:beta}\beta_k(\prem, \hypo) &= \prem c_1\dots c_k \hypo
\end{align}
Note that $\alpha$ divides its $k$ tokens into three parts as equally as possible
where any remaining tokens go between $\prem$ and $\hypo$
if $k$ is not a multiple of $3$.
In particular, this means that the same templates are produced by $\alpha$ and $\beta$ for $k\leq 2$.
We chose this behavior as a generalization of the standard approach to fine-tuning a PLM for sequence classification (such as NLI)
where there is only one special token and it separates the two input sequences.
The template produced by $\alpha_1$ and $\beta_1$ is very similar to this.
A major difference is that the embeddings for $\contokens$ tokens are randomly initialized
whereas the standard separator token has a pretrained embedding.


\para{Pattern-based classifier}
Given $\gamma \in \eset{\alpha, \beta}$,
we estimate the probability distribution $\cprob{\hat{y}}{\prem, \hypo}$
with a linear classifier on top of the pooled sequence representation produced by the PLM $\lm$:
\begin{equation}
	\cprob{\hat{y}}{\prem, \hypo} = \sigma(\lm(\gamma(\prem, \hypo))W + b)
\end{equation}
where $W\in\mathbb{R}^{d\times 2}, b\in\mathbb{R}^{2}$ are learnable parameters,
$\sigma$ is the softmax function,
and applying $\lm$ means encoding the whole input sequence in a single $d$-dimensional vector
according to the specifics of the PLM.
For BERT \citep{devlin-etal-2019-bert} and its successor RoBERTa \citep{liu19},
this implies a dense pooler layer with tanh activation over the contextualized token embeddings
and picking the first of these embeddings (i.e., $\mathtt{[CLS]}$ for BERT and $\mathtt{\langle s\rangle}$ for RoBERTa).\footnote{Cf.\ Jacob Devlin's comment on issue 43 in the official BERT repository on GitHub, \url{https://github.com/google-research/bert/issues/43}.}
For training, we apply dropout with a probability of $0.1$ to the output of $\lm(\cdot)$.
	
\para{Inference with multiple patterns}
Previous work \citep{bouraoui20,schmitt-schutze-2021-language} combined multiple patterns with the intuition
that different NL patterns can capture different aspects of the task.
This intuition makes also sense for \conan{}.
We conjecture
that an efficient use of the model parameters requires different continuous patterns to learn different representations,
which can detect different types of entailment.
Following the aforementioned work,
we form our final score $s$ by combining the probability estimates from different patterns $\Gamma$
by comparing the maximum probability for the two classes $0,1$ over all patterns:
\begin{align}
	\nonumber m_{1}(\prem, \hypo) &= \max_{\gamma\in\Gamma} \cprob{\hat{y} = 1}{\gamma(\prem, \hypo)} \\
	\nonumber m_{0}(\prem, \hypo) &= \max_{\gamma\in\Gamma} \cprob{\hat{y} = 0}{\gamma(\prem, \hypo)} \\
	s(\prem, \hypo) &= m_{1}(\prem, \hypo) - m_{0}(\prem, \hypo)
\end{align}

In conclusion, a \conan{} model $\gamma^{n}_{k}$ is characterized by three factors:
\begin{enumerate*}[label=(\roman{*})]
	\item The type of pattern $\gamma\in\eset{\alpha, \beta}$,
	\item the number of patterns $n\in\mathbb{N}$, and
	\item the number of tokens $k\in\mathbb{N}$ per pattern.
\end{enumerate*}

\para{Training}
While multiple patterns are combined for decision finding during inference,
we treat all patterns separately during training -- as did previous work \citep{schmitt-schutze-2021-language}.
So, given a set of patterns $\Gamma$, we minimize the negative log-likelihood of the training data $\mathcal{T}$, i.e.,
\begin{align}
	\sum_{(\prem, \hypo, y)\in\mathcal{T}} \sum_{\gamma\in\Gamma} -\log(\cprob{\hat{y} = y}{\gamma(\prem, \hypo)})
\end{align}
In practice, we apply mini-batching to both $\mathcal{T}$ and $\Gamma$ and thus compute this loss only for a fraction of the available training data and patterns at a time.
In this case, we normalize the loss by averaging over the training samples and patterns in the mini-batch.


\begin{table}[t]
	\centering
	\begin{tabular}{lrrrr}
		\toprule
		&train & dev & test & total\\
		\midrule
		SherLIiC &797&201&2,990 & 3,988\\
		Levy/Holt &4,388&1,098&12,921 & 18,407\\
		\bottomrule
	\end{tabular}
	\caption{Number of labeled instances per data split.}
	\label{tab:data}
\end{table}

\begin{figure}[t]
	\begin{subfigure}{\linewidth}
		\includegraphics[width=\linewidth]{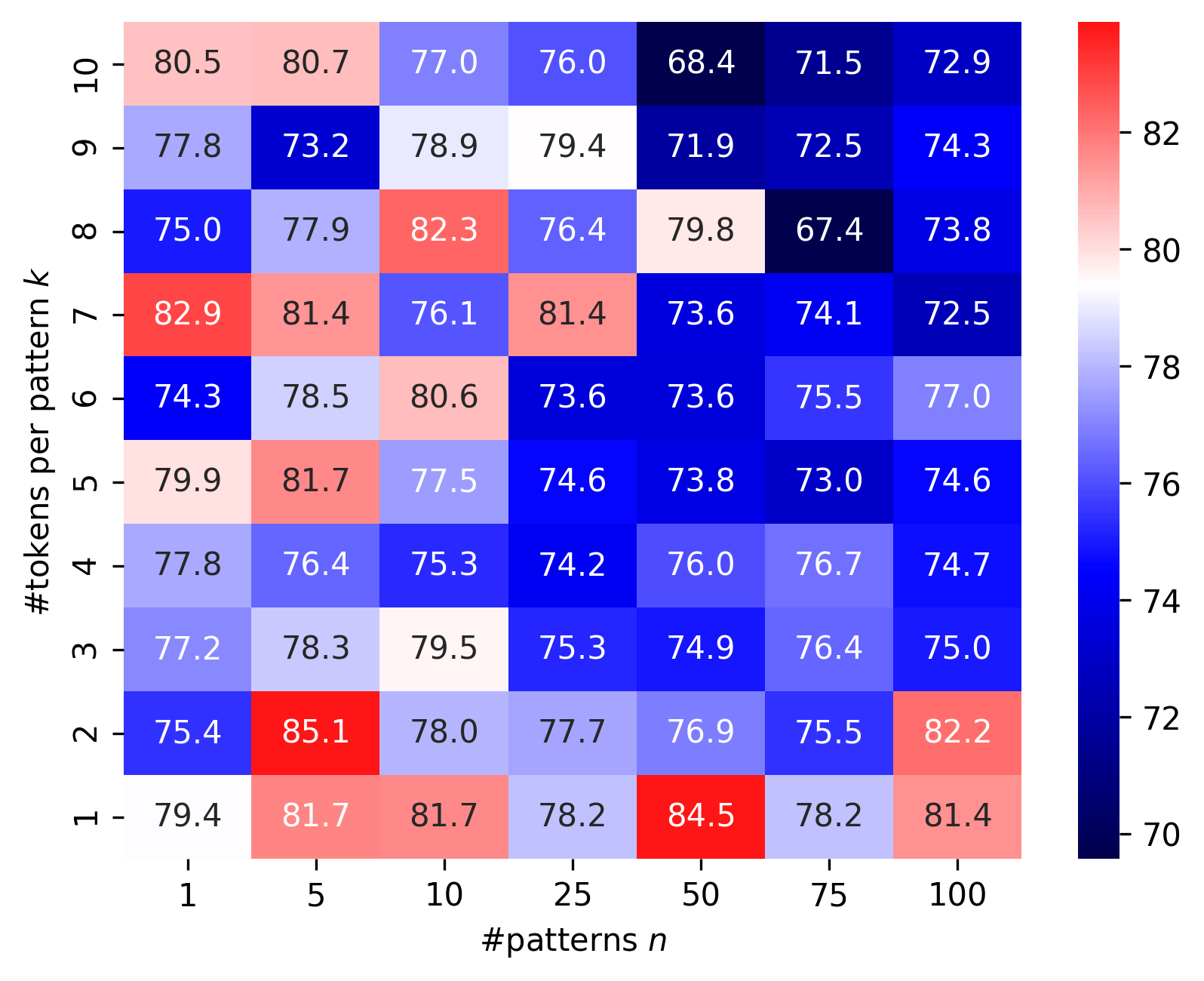}
	\end{subfigure}
	\begin{subfigure}{\linewidth}
		\includegraphics[width=\linewidth]{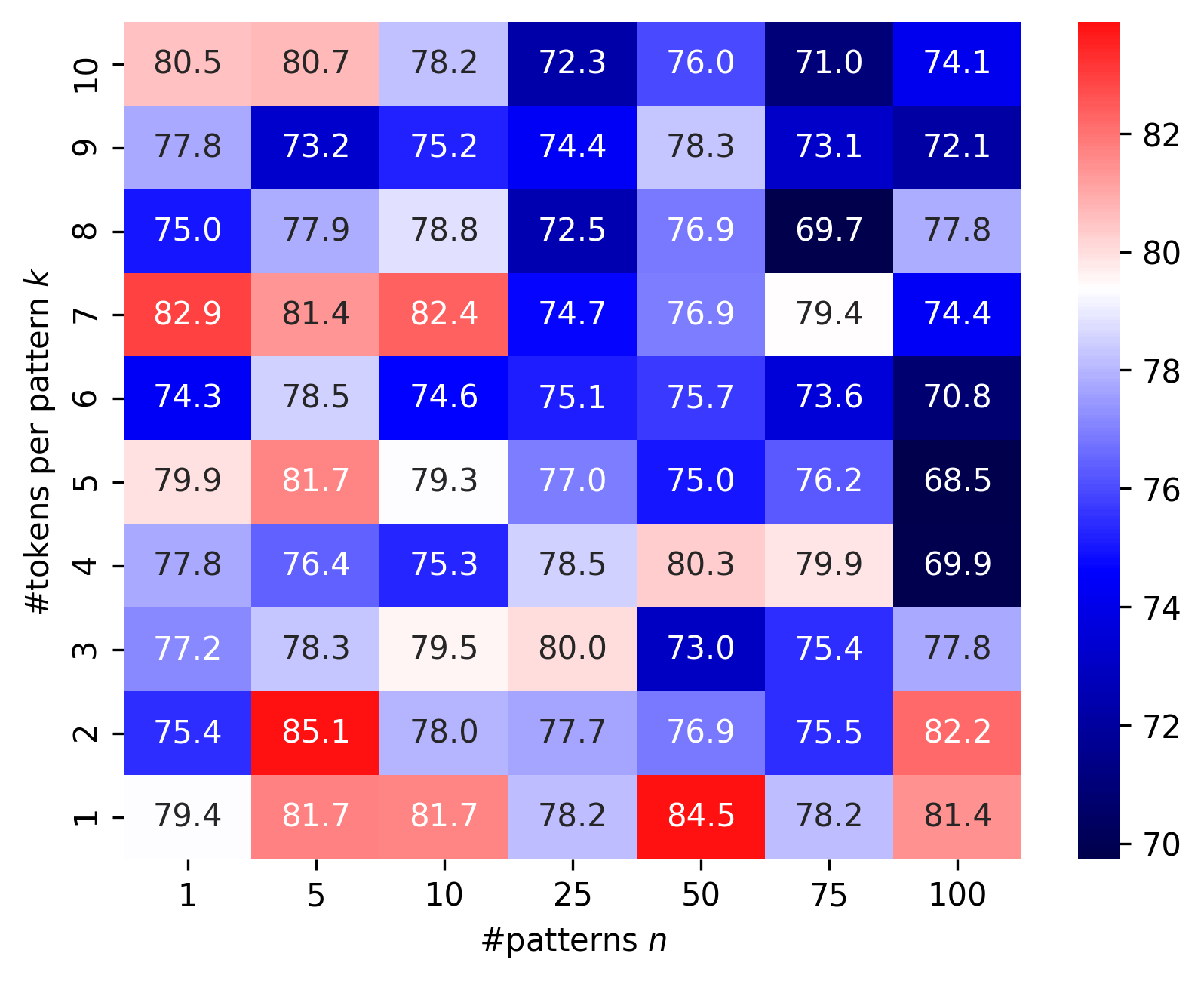}
	\end{subfigure}
	\caption{\auc{} on SherLIiC dev for different \conan{} models; top = $\alpha$, bottom = $\beta$; white/red/blue = similar to/better than/worse than $n=k=1$. Note that $\alpha_k = \beta_k$ for $k \leq 2$.}
	\label{fig:dev_auc}
\end{figure}

\section{Experiments}

We conduct experiments on two established LIiC benchmarks,
SherLIiC \citep{schmitt-schutze-2019-sherliic} and Levy/Holt \citep{levy-dagan-2016-annotating,holt18},
using the data splits as defined in \citep{schmitt-schutze-2021-language} for comparison.
%
Both benchmarks contain a majority of negative examples (SherLIiC: 67\%{}, Levy/Holt: 81\%{}), making the detection of the entailment (i.e., the minority) class a particular challenge.
See \cref{tab:data} for dataset and split sizes.
Note that Levy/Holt is nearly 5 times bigger than SherLIiC
and still has less than 5k train samples.

Following \citep{schmitt-schutze-2021-language},
we use RoBERTa as underlying PLM and also use the same hyperparameters whenever possible for comparison.
Also following \citet{schmitt-schutze-2021-language}, we
instantiate the typed placeholders $A,B$ in SherLIiC with
Freebase \citep{bollacker08} entities,
making sure that $A$ and $B$ are not assigned the same entity.
See \cref{app:hparam} for full training details.

We evaluate model performance with two metrics:
\begin{enumerate*}[label=(\roman{*})]
	\item The area under the precision-recall curve for precision values $\geq$ 0.5 (\auc) as threshold-less metric using only the score $s$ defined in the previous section and
	\item the F1 score of actual classification decisions after tuning a decision threshold $\vartheta$ on the respective dev portion of the data.
\end{enumerate*}
Our implementation is based on \citep{huggingface}.

\section{Results}
\label{sec:results}

\para{Choosing $n$ and $k$}
The number $n$ of patterns and the number $k$ of $\contokens$ tokens per pattern
are essential hyperparameters of a \conan{} model.

\cref{fig:dev_auc} shows the impact on performance (measured in \auc{}) on SherLIiC dev for different $n$-$k$-combinations.
We observe that using too many, too long patterns harms performance w.r.t.\ the base case $n=k=1$.
Best results are obtained with either a small number of patterns or tokens or both.
Comparing the $\alpha$ and $\beta$ settings,
we notice that, rounded to one decimal, they produce identical results for both $n=1$ and $n=5$ patterns,
suggesting that the particular position of the $\contokens$ tokens does not matter much in these settings for SherLIiC.
Even with $n=10$ patterns,
the two methods only begin to differ with $k \geq 5$ tokens per pattern.
Evaluation on the Levy/Holt data (see \cref{fig:dev_auc_levy} in \cref{app:more_dev}) shows more variation between $\alpha$ and $\beta$ but, otherwise, confirms the trend
that small $n$ and $k$ yield better performance.
%

Our results offer an explanation for the empirical finding in \citep{schmitt-schutze-2021-language}
that patterns retrieved from a corpus lead to worse performance than handcrafted ones
because the latter are generally shorter.
\conan{} models do not only yield better performance,
they also provide an automatic way to test pattern properties,
such as length,
w.r.t.\ effect on performance for a given task.

\begin{table}[t]
	\centering
	\small
	\begin{tabular}{l@{\, }lrrrr}
		\toprule
		&&\auc&P&R&F1\\
		\midrule
		\multicolumn{6}{l}{{\tiny RoBERTa-base}}\\
		\textsc{manpat}$^\Phi$&{\tiny (baseline)}&69.2& 62.0& 81.2& 70.3\\
		\textsc{autpat}$_{15}^\Phi$&{\tiny (baseline)}&73.1&63.0&77.4&69.4\\
		\conan{}$^5_2$&{\tiny ($\vartheta=-0.0768$)}&73.2&65.8&\textbf{84.1}&\textbf{73.8}\\
		\conan{}$^{50}_1$&{\tiny ($\vartheta=-0.0108$)}&\textbf{74.0}&\textbf{70.1}&78.0&\textbf{73.8}\\
		\conan{}-$\alpha^1_7$&{\tiny ($\vartheta=-0.1875$)}&66.9&68.7&68.4&68.5\\
		\conan{}-$\beta^1_7$&{\tiny ($\vartheta=-0.1875$)}&66.9&68.7&68.4&68.5\\
		\midrule
		\multicolumn{6}{l}{{\tiny RoBERTa-large}}\\
		\textsc{nli} &{\tiny (baseline)}& 68.3 & 60.5 & 85.5 & 70.9 \\
		$\textsc{manpat}^{\Phi\Psi}$ &{\tiny (baseline)}& 74.4 & 66.0 & 80.8 & 72.6 \\
		\conan{}$^5_2$&{\tiny ($\vartheta=-0.6838$)}&\textbf{75.9}&\textbf{67.9}&81.1&\textbf{73.9}\\
		\conan{}$^{50}_1$&{\tiny ($\vartheta=-0.9999$)}&73.9&64.8&81.5&72.2\\
		\conan{}-$\alpha^1_7$&{\tiny ($\vartheta=-0.8797$)}&62.6&63.1&76.4&69.1\\
		\conan{}-$\beta^1_7$&{\tiny ($\vartheta=-0.9556$)}&68.6&60.8&\textbf{86.0}&71.2\\
		\bottomrule
	\end{tabular}
	\caption{SherLIiC test. \auc{} denotes the area under the precision-recall curve for precision $\geq$ 0.5. All results in \%{}. Bold means best result per column and block. All baselines from \citep{schmitt-schutze-2021-language}. For $k\leq 2$, we simply write \conan{}$^n_k$ because $\alpha = \beta$.}
	\label{tab:sherliic}
\end{table}
\begin{table}[t]
	\centering
	\small
	\begin{tabular}{l@{\, }lrrrr}
		\toprule
		&&\auc&P&R&F1\\
		\midrule
		\multicolumn{6}{l}{{\tiny RoBERTa-base}}\\
		\textsc{nli} &{\tiny (baseline)}&72.6&68.7&75.3&71.9\\
		$\textsc{manpat}^{\Phi\Psi}$ &{\tiny (baseline)}&76.9&78.7&66.4&72.0\\
		\conan{}$^5_2$&{\tiny ($\vartheta=-0.4809$)}&\textbf{77.6}&\textbf{79.1}&67.5&\textbf{72.8}\\
		\conan{}$^{5}_1$&{\tiny ($\vartheta=-0.9003$)}&74.9&67.1&\textbf{77.5}&71.9\\
		\conan{}-$\alpha^5_3$&{\tiny ($\vartheta=-0.8985$)}&73.6&75.0&66.7&70.6\\
		\conan{}-$\beta^5_3$&{\tiny ($\vartheta=-0.9289$)}&76.4&76.6&69.2&72.7\\
		\midrule
		\multicolumn{6}{l}{{\tiny RoBERTa-large}}\\
		$\textsc{manpat}^{\Phi\Psi}$ &{\tiny (baseline)}&83.9&84.8&70.1&76.7\\
		$\textsc{manpat}^{\Phi}$ &{\tiny (baseline)}&77.8&67.9&\textbf{81.5}&74.1\\
		\conan{}$^5_2$&{\tiny ($\vartheta=-0.1315$)}&\textbf{85.9}&81.7&78.1&\textbf{79.9}\\
		\conan{}$^{5}_1$&{\tiny ($\vartheta=-0.9750$)}&85.2&77.2&80.3&78.7\\
		\conan{}-$\alpha^5_3$&{\tiny ($\vartheta=-0.9212$)}&84.4&\textbf{82.2}&74.9&78.4\\
		\conan{}-$\beta^5_3$&{\tiny ($\vartheta=-0.9585$)}&85.3&78.8&77.3&78.0\\
		\bottomrule
	\end{tabular}
	\caption{Levy/Holt test. All baselines from \citep{schmitt-schutze-2021-language}. See \cref{tab:sherliic} for table format.}
	\label{tab:levyholt}
\end{table}

\para{Test performance}
%
On both SherLIiC (\cref{tab:sherliic}) and Levy/Holt (\cref{tab:levyholt}),
and across model sizes (base and large),
\conan{}$^5_2$ (using either $\alpha$ or $\beta$ because they are identical for $k=2$) outperforms all other models including the previous state of the art by \citet{schmitt-schutze-2021-language},
who fine-tune RoBERTa both without patterns (\textsc{nli}) and using handcrafted (\textsc{manpat}) or automatically retrieved corpus patterns (\textsc{autpat}).
We report their two best systems for each benchmark.

We take the performance increase with continuous patterns as a clear indicator that the flexibility offered by separating pattern tokens from the rest of the vocabulary allows RoBERTa to better adapt to the task-specific data
even with only few labeled training instances in the challenging LIiC task.

\section{Analysis and Discussion}
\label{sec:analysis}

\para{Nearest neighbors}
To further investigate how RoBERTa makes use of the flexibility of $\contokens$ tokens,
we compute their nearest neighbors in the space of original vocabulary tokens based on cosine similarity for our models in \cref{tab:sherliic,tab:levyholt}.
We always find the $\contokens$ tokens to be very dissimilar from any token in the original vocabulary,
the highest cosine similarity being $0.15$.
And even among themselves, $\contokens$ tokens are very dissimilar, nearly orthogonal,
with $0.08$ being the highest cosine similarity here.
RoBERTa seems to indeed take full advantage of the increased flexibility to put the $\contokens$ tokens anywhere in the embedding space.
This further backs our hypothesis that the increased flexibility is beneficial for performance.

\para{Influence of additional parameters}
One might argue that the vocabulary extension and the resulting new randomly initialized token embeddings lead to an unfair advantage for \conan{} models because the parameter count increases.
While more parameters do generally lead to increased model capacity,
the number of new parameters is so small compared to the total number of parameters in RoBERTa that we consider it improbable
that the new parameters are alone responsible for the improved performance.
Of all models in \cref{tab:sherliic,tab:levyholt},
\conan{}$^{50}_1$ introduces the most additional model parameters,
i.e., $1\cdot 50\cdot 768 = 38400$ for RoBERTa-base.
Given that even the smaller RoBERTa-base model still has a total of 125M parameters,
the relative parameter increase is maximally 0.03\%{}, which, we argue, is negligible.

\begin{table}[t]
	\centering
	\small
	\begin{tabular}{lrrrr}
		\toprule
		&\auc{}&P&R&F1\\
		\midrule
		\multicolumn{5}{l}{{\tiny SherLIiC train $\to$ Levy/Holt test}}\\
		Schm\&{}Schü (2021) {\tiny (base)}&38.4&\textbf{52.7}&57.1&54.8\\
		Schm\&{}Schü (2021) {\tiny (large)}&\textbf{70.4}&39.6&95.3&\textbf{56.0}\\
		\conan{}$^5_2$ {\tiny (base)}&54.3&38.2&89.8&53.5\\
		\conan{}$^5_2$ {\tiny (large)}&61.5&34.5&\textbf{96.3}&50.8\\
		\midrule
		\multicolumn{5}{l}{{\tiny Levy/Holt train $\to$ SherLIiC test}}\\
		Schm\&{}Schü (2021) {\tiny (base)}&63.3&62.8&68.4&65.5\\
		Schm\&{}Schü (2021) {\tiny (large)}&62.1&68.1&57.3&62.3\\
		\conan{}$^5_2$ {\tiny (base)}&64.9&63.4&68.9&66.1\\
		\conan{}$^5_2$ {\tiny (large)}&\textbf{70.1}&\textbf{69.0}&\textbf{69.5}&\textbf{69.2}\\
		\bottomrule
	\end{tabular}
	\caption{Transfer experiments ($\vartheta=0$). Best models from \citep{schmitt-schutze-2021-language} according to F1 score.}
	\label{tab:transfer}
\end{table}

\para{Transfer between datasets}
The experiments summarized in \cref{tab:transfer} investigate the hypothesis
that \conan{}'s better adaptation to the fine-tuning data might worsen its generalization abilities to other LIiC benchmarks.
For this, we train our best model \conan{}$^5_2$ on SherLIiC to test it on Levy/Holt and vice versa.
In this scenario, we assume that the target dataset is not available at all.
So there is no way to adapt to a slightly different domain other than learning general LIiC reasoning.
We thus set $\vartheta=0$ in these experiments.

We find that with the very few train samples in SherLIiC the risk of overfitting to SherLIiC is indeed higher.
When trained on Levy/Holt with around 4.4k train samples, however, \conan{} clearly improves generalization to the SherLIiC domain.

\section{Related Work}
\label{sec:related_work}

\para{PLMs and text patterns}
GPT-2 \citep{radford19} made the idea popular that a PLM can perform tasks without access to any training data
when prompted with the right NL task instructions.
With GPT-3, \citet{brown20} adapted this idea to few-shot settings where the task prompt is extended by a few training samples.
While this kind of few-shot adaptation with a frozen PLM only works with very big models,
\citet{schick-schutze-2021-it} achieve similar performance with smaller models by fine-tuning the PLM on the available training data and putting them into NL templates.
Recently, \citet{schmitt-schutze-2021-language} investigated the use of PLMs for LIiC.
Compared to a standard sequence classification fine-tuning approach,
they were able to improve the PLM RoBERTa's performance by putting an entailment candidate into textual contexts that only make sense for either a valid or invalid example.
Patterns like ``$y$ because $x$.'' (valid) or ``It does not mean that $y$ just because $x$.'' (invalid) make intuitive sense to humans and outperform standard RoBERTa on LIiC. 

A large problem with all these approaches, however, is to find well-functioning patterns,
for which numerous solutions have been proposed \citep{shin20,haviv-etal-2021-bertese,bouraoui20,jiang-etal-2020-know,gao20,reynolds21}.
We argue that it is not optimal to constrain pattern search to the space of NL sequences
if the primary goal is better task performance,
and therefore abandon this constraint.

\para{PLMs and continuous patterns}
\citet{li21} and \citet{hambardzumyan21} contemporaneously introduced the idea of mixing the input token embeddings of a PLM with other continuous vectors that do not correspond to vocabulary elements.
In the spirit of GPT-2 (see above), they keep the PLM's parameters frozen and only fine-tune the embeddings of the ``virtual tokens'' to the target task.
While this line of research offers certain appeals of its own, e.g., reusability of the frozen PLM weights,
this is not the focus of our work.
In pursuit of the best possible performance,
we instead compare the use of continuous vs.\ NL patterns in the process of fine-tuning all PLM parameters
and find that even carefully chosen NL patterns can be outperformed by our automatically learned ones.

Contemporaneously to our work,
\citet{liu21} fine-tune entire PLMs with continuous patterns for SuperGLUE \citep{wang19}.
Besides reformulating the SuperGLUE tasks as cloze tasks, while we keep formalizing our task as classification,
\citet{liu21} also add more complexity by computing the continuous token representations with an LSTM \citep{hochreiter97}
and adding certain ``anchor tokens'', such as a question mark, at manually chosen places.
\conan{} does not use any manual pattern design and embeds continuous tokens with a simple lookup table.

Another contemporaneous work by \citet{lester21} tests the influence of model size on the performance of a frozen PLM with trained continuous prompts.
Their prompt ensembling is akin to our combining multiple patterns during inference (cf.\ \cref{sec:conan}).
The key difference is that, instead of making predictions with different patterns and taking the majority vote,
we rather compare the scores for different patterns to make our prediction.

\section{Conclusion}
We presented \conan{}, a method that improves fine-tuning performance of a PLM with continuous patterns.
\conan{} does not depend on any manual pattern design and is
efficient as the shortest possible patterns with good performance can be found automatically.
It provides an automatic way of systematically testing structural properties of patterns, such as length, w.r.t.\ performance changes.
In our experiments on two established LIiC benchmarks,
\conan{} outperforms previous work using NL patterns and sets a new state of the art.


\section*{Acknowledgments}
We gratefully acknowledge a Ph.D. scholarship
awarded to the first author by the German Academic Scholarship Foundation (Studienstiftung des
deutschen Volkes). This work was supported by the
BMBF as part of the project MLWin (01IS18050).

\bibliography{anthology,custom}
\bibliographystyle{acl_natbib}

\appendix
\section{Training Details}
\label{app:hparam}

We train our model for 5 epochs on a single GeForce RTX 2080 Ti GPU,
with Adam \citep{kingma15} and a mini-batch size of 10 (resp.\ 2) training instances for RoBERTa-base (resp.\ -large) and mini-batches of 5 patterns.  
We adopt well-functioning values from \citep{schmitt-schutze-2021-language}
for all non-\conan{}-specific hyperparameters, i.e., learning rate $\mathit{lr}$, weight decay $\lambda$, and accumulated batches $c$ before a gradient update:
Concretely, we set $\mathit{lr} = 2.28\cdot 10^{-5}, \lambda = 6.52\cdot10^{-2}, c=2$ for evaluating RoBERTa-base on SherLIiC,
$\mathit{lr} = 1.29\cdot 10^{-5}, \lambda = 2.49\cdot10^{-4}, c=3$ for RoBERTa-large on SherLIiC,
$\mathit{lr} = 2.72\cdot 10^{-5}, \lambda = 1.43\cdot10^{-3}, c=1$ for RoBERTa-base on Levy/Holt,
and $\mathit{lr} = 4.55\cdot 10^{-6}, \lambda = 3.90\cdot10^{-4}, c=2$ for RoBERTa-large on Levy/Holt.

\section{More Dev Results}
\label{app:more_dev}

\begin{figure}[t]
	\begin{subfigure}{\linewidth}
		\includegraphics[width=\linewidth]{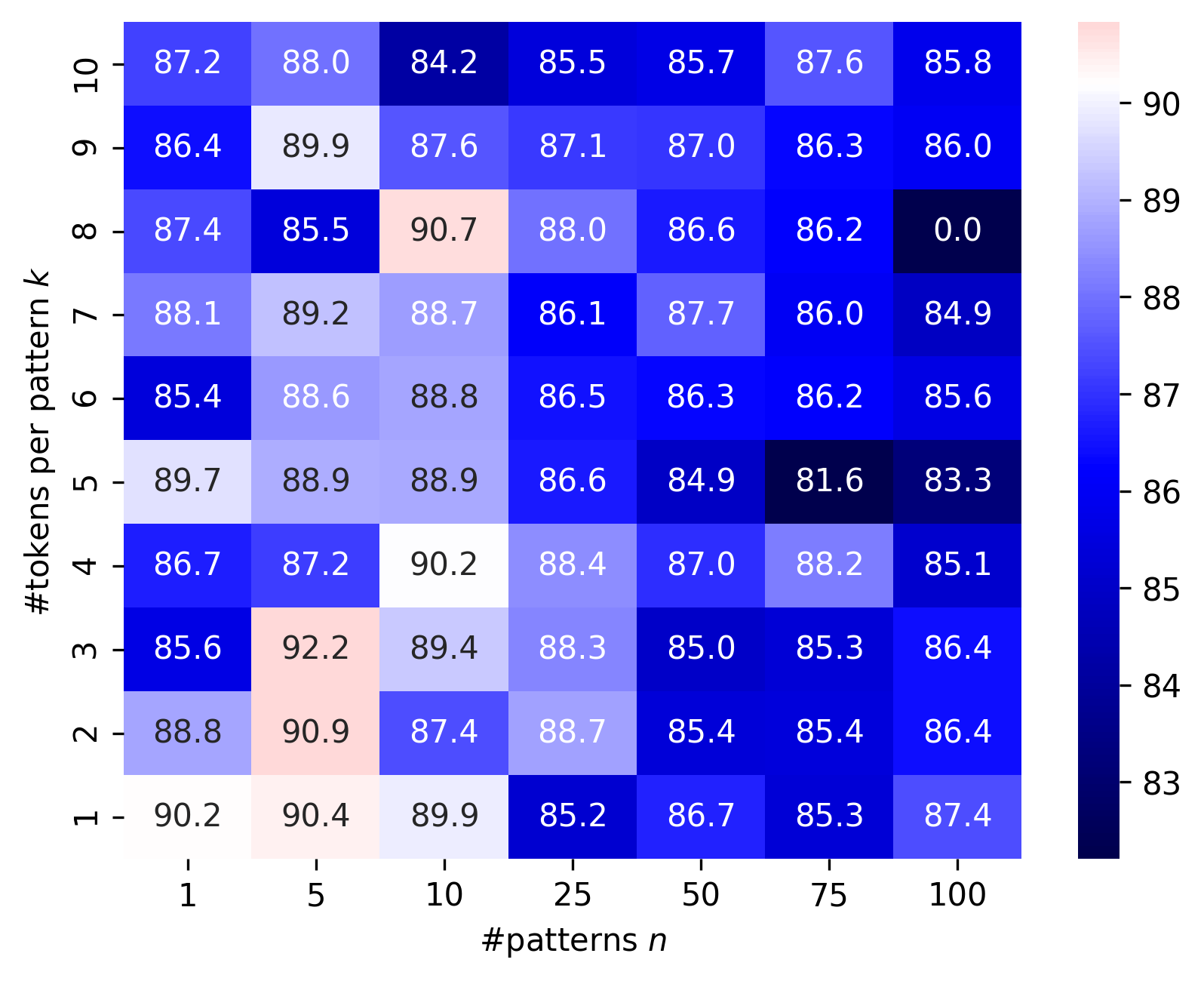}
	\end{subfigure}
	\begin{subfigure}{\linewidth}
		\includegraphics[width=\linewidth]{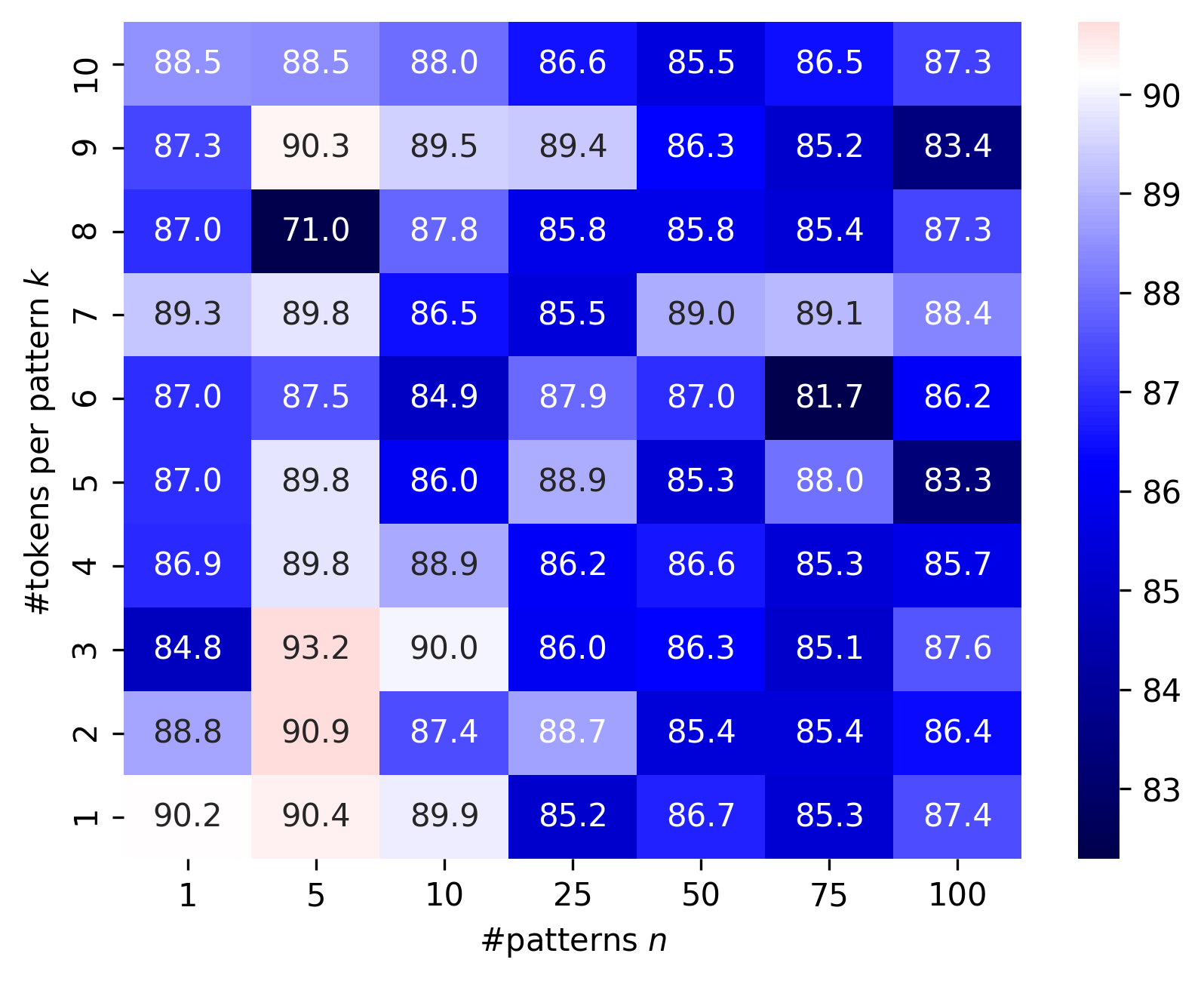}
	\end{subfigure}
	\caption{\auc{} on Levy/Holt dev for different \conan{} models. Same format as \cref{fig:dev_auc}.}
	\label{fig:dev_auc_levy}
\end{figure}

See \cref{fig:dev_auc_levy} for evaluation results on Levy/Holt dev.

%

\end{document}